\documentclass[english]{IEEEtran}
\usepackage[T1]{fontenc}
\usepackage[latin9]{inputenc}
\usepackage{color}
\usepackage{babel}
\usepackage{float}
\usepackage{graphicx}
\usepackage[unicode=true,pdfusetitle,
 bookmarks=true,bookmarksnumbered=false,bookmarksopen=false,
 breaklinks=false,pdfborder={0 0 1},backref=false,colorlinks=false]
 {hyperref}

\makeatletter
\usepackage[dvipsnames]{xcolor}

\makeatother

\usepackage{listings}

\begin{document}
\title{An Image Labeling Tool and Agricultural Dataset for Deep Learning}
\author{Patrick Wspanialy, Justin Brooks, Medhat Moussa}
\maketitle
\begin{abstract}
We introduce a labeling tool and dataset aimed to facilitate computer
vision research in agriculture. The annotation tool introduces novel
methods for labeling with a variety of manual, semi-automatic, and
fully-automatic tools. The dataset includes original images collected
from commercial greenhouses, images from PlantVillage, and images
from Google Images. Images were annotated with segmentations for foreground
leaf, fruit, and stem instances, and diseased leaf area. Labels were
in an extended COCO format. In total the dataset contained 10k tomatoes,
7k leaves, 2k stems, and 2k diseased leaf annotations. 
\end{abstract}

\section{Introduction}

Over the past decade computer vision research has shifted away from
the design of manually defined feature descriptors and towards learning
directly from data. As a result of innovations in deep learning, models
have achieved super-human levels of performance on a variety of vision
tasks. \cite{heDeepResidualLearning2015,estevaDermatologistlevelClassificationSkin2017,DeepLungDeep3D}
Convolutional neural networks use layers of artificial neurons to
increasingly learn higher levels of object representations, beginning
with simple lines and corners and towards complete objects such as
fruits and flowers. 

A major challenge to computer vision using supervised deep learning
is the large amount and variety of annotated images required to automatically
discover salient features for object recognition, detection, and segmentation.
Due to this challenge, pioneering deep learning studies \cite{dengImageNetLargescaleHierarchical2009,krizhevskyImageNetClassificationDeep2012a}
focused on classification tasks and used images with single prominent
objects collected using web searches of associated filenames and metadata.
This approach simplified the annotation process, as only short text
descriptions were necessary. As more advanced forms of scene understanding
were developed, the need arose for more sophisticated annotation tools. 

The Common Objects in COntext (COCO) \cite{linMicrosoftCOCOCommon2014b}
dataset introduced images with detailed annotations of objects in
the context of their natural surroundings with the goal of advancing
computer vision beyond image classification. In total there were 2.5
million instances from 80 categories in 328k images labeled. The annotations
in COCO included bounding boxes and pixel-level segmentation of each
object instance. Images on average contained 7 instances and took
1 minute and 19 seconds to annotate. Annotations were created using
the \emph{opensurface-segmentation-ui }annotation tool \cite{bellOpenSurfacesRichlyAnnotated2013},
which only drew shapes using vertex-clicking. 

Precision agriculture is a quickly growing market segment \cite{PrecisionFarmingMarket},
with a substantial research effort focusing on computer vision. \cite{capDeepLearningApproach2018,grinblatDeepLearningPlant2016,mohantyUsingDeepLearning2016b}
The agriculture community has a considerably smaller collection of
publicly available data to use in computer vision experimentation
compared with more generalized computer vision research. The limited
access to public data presence barriers to creating comparative studies,
peer review of papers, and measuring the progress the community is
making over time. Agriculture studies have traditionally collected
their own data and used it to develop and test new algorithms. As
a result, datasets are often relatively small and not made public.
In circumstances where data is released, there is no common format
to follow which complicates testing and advancement of methods by
third parties.

There have been some efforts to collect large amounts of public agricultural
image data, most notably by PlantVillage \cite{hughesOpenAccessRepository2015a}
in 2015. The dataset contained over 50k leaf images organized by experts
into plant and disease type categories. The PlantVillage project is
focused on helping small growers better manage crops through the use
of technology to help increase the global food supply. One of the
most pressing concerns for small growers is disease, as it is frequently
the cause of a total-loss harvest. The original dataset contained
high resolution imagery of the diseased leaves, and was announced
as a public dataset. However, as project focus shifted to other research
areas access to the data was removed. A low resolution version of
the dataset is still currently available in an online code repository
for a related research publication. \cite{mohantyUsingDeepLearning2016b} 

\subsection{Contribution}

We introduce a annotation tool and dataset aimed at facilitating computer
vision research in agriculture. The annotation tool provides novel
methods for labeling, with a variety of manual, semi-automatic, and
fully-automatic tools. In order to facilitate future integration and
expansion of the dataset, we also introduce a dataset format compatible
with many existing computer vision tools in the community. The dataset
builds on images from PlantVillage by adding pixel-level segmentations
of diseased areas, and includes images compiled from Google Image
searches, and from originally collected images of commercial greenhouse
plants.

\section{Labeling techniques}

The annotation tool introduced here builds on previous techniques
used to label images and incorporates several novel annotation workflows.
As agricultural images often have many more individual objects and
with more intricate shapes then those in many general datasets, this
tool focused on increasing the efficiency of labeling these types
of images.

The labeling tool was designed to be used with a keyboard and mouse,
and includes several keyboard shortcuts and mouse gestures to help
improve annotation efficiency. 

The annotator's labeling techniques could be used independently or
in combination to label each object. Annotated objects can be composed
of several disjoint segments to accommodate occlusion. Besides a main
label, each object can be associated with an unlimited amount of metadata
to provide additional context. For example, a fruit can have meta-data
describing its ripeness and associated plant ID. 

The following is a description of each annotation technique.

\subsection{Free-form polygons}

The vertex-clicking technique used in previous system focused on manufactured
objects with simple and sharp corners. When used to annotate natural
objects with curved edges, such as those found in typical agricultural
scenes, the resulting shapes exhibited low fidelity boundaries. Higher
quality boundaries were possible by adding many intermediate points
along a curve but at the cost of significantly increasing annotation
time and effort. 

The free-form polygon tool allows users to click-drag-release around
objects to annotate them, while the system automatically assigns points
along the boundary. Polygon precision and the minimum distance between
the first and last point to automatically complete the shape can be
adjusted using a parameter in the user interface. After the polygon
is generated, the select tool can be used to individually move each
vertex independently to fine tune the final shape.

This is the tool which was most often used during annotation of the
included dataset. 

\subsection{Flood fill}

The flood fill tool selects areas of uniform color and automatically
generates an enclosing polygon with adjustable vertices. 

Two selectable parameters adjust the color threshold and the size
of a gaussian blur used to determine the enclosing area. Areas can
be removed by using the shift modifier key and clicking inside an
existing polygon. 

This tool works best for large intricate areas of uniform color, and
can generate complex polygons with a single click. 

\subsection{Brush and eraser}

The brush tool allowed adding annotation by painting using a technique
similar to that of familiar drawing software. The size of the brush
could be adjusted using a parameter or keyboard shortcut, and areas
were selected using a click-drag-release technique. The resulting
area was automatically converted into a detailed polygon which could
be further adjusted by moving its vertices. 

The eraser tool functioned in a similar way to the brush but removed
annotated areas instead of adding them. The eraser was not limited
to erasing areas created using the brush tool, and could be used to
fine-tune any generated polygon. 

\subsection{Key points}

Key points are used for describing the spatial relationships between
objects or parts of objects. It is most notably used to characterize
the human skeleton for analysis and tracking. As a result of this
primary application, the original key point system assumed a fixed
number of points with unchanging relationships (toe bone connected
to the foot bone, etc). 

When key points are used to represent the structure of plant stems,
key points structure must become dynamic to properly characterize
nodes as the plant grows. 

The key point annotation tool provides the functionality to describe
important specific locations on objects and their relationships to
one another. Key points can be used on their own or associated with
a segmented annotation. Each point has its own ID, label, list of
connected points, and a visibility property, indicating whether the
object the point represents is shown or is obscured in some way. Points
can be dragged to new locations after being placed, and relationships
are created by clicking on corresponding points. 

\subsection{Deep Extreme Cut}

Deep extreme cut (DEXTR) \cite{Man+18} is semi-automated tool for
object segmentation, requiring four only points to be defined by the
user to produce precise results. This manual labeling technique produces
bounding boxes using the extreme left, right, top, and bottom points
of an object instead of the traditional two-point approach. Although
twice the input points, the four-point approach was shown to significantly
reduce labeling time, as two-point bounding box starting coordinates
are often difficult to estimate and require additional adjustment. 

The four points are used to guide a convolutional neural network (CNN)
which was previously trained on manually annotated examples. The performance
measured using intersection over union on unseen categories during
training was approximately 80\%.

The padding surrounding the bounding box created by the four points
can be adjusted through a UI parameter to reduce the effect of errors
when defining the extreme points. The resulting annotation is converted
into a polygon which can be further manually refined by either moving
individual vertices or by using one of the other annotation tools. 

\subsection{Mask R-CNN}

Mask R-CNN \cite{heMaskRCNN2017a} is a deep learning model using
convolutional neural networks which learn how to segment every object
instance in an image. While this type of model may be the final objective
of annotating a dataset, an intermediate version can be used to pre-annotate
a portion of image objects and improve overall efficiency. 

After an initial fraction of a dataset is annotated, a pre-trained
Mask R-CNN model is fine-tined on the annotated object classes. The
training input to a Mask R-CNN model are the annotations made using
the techniques previously described. 

Once a model is trained the auto-annotate button in the user interface
is activated. When pressed, the currently selected image is sent to
the model's API, which returns a list of objects and their segmentations
as polygons. Each object's polygon is added to the image and can be
manipulated using the other techniques to correct errors.

As a larger fraction of the dataset is annotated, the Mask R-CNN model
performance improves, further increasing annotation efficiency by
requiring fewer corrections. 

\section{Annotation pipeline }

\subsection{Collect images}

When collecting images for use with deep learning it is important
to consider the intended application of the models. The features learned
during the training process are exclusivity derived from the data
provided, and if the dataset is biased in an unexpected way, model
results can be misleading. 

In order to reduce bias between and within object classes, the number
of samples between classes should equal and each class should have
a minimum of 1000. Variations in the appearance of samples collected
should reflect the variation expected during application. In general
more samples are required with a less constrained environments. 

Another important step to building a dataset is to define a clear
problem definition. For example, if locating a specific type of fruit
within an image a single class may be sufficient. When grading fruit
for ripeness, the dataset should have labels for each stage relevant
to the growing and harvesting processes. If predicting when a fruit
will reach a ripeness stage, additional metadata indicating a relative
measure of time between samples will be required. 

The selection of camera configuration and lighting requires carefully
attention, as they may introduce sources of bias if they are consistent
within classes but different between them. This can become especially
prominent when inter-class differences are small, as with fruit ripeness
or leaf disease applications. Models may learn to discriminate samples
based on inconspicuous features such as image noise or white balance.

The ``generate'' feature of the annotation tool downloads a set
of images from Google associated with a selected keyword, and can
be used to create a preliminary dataset or to augment an existing
one with a wider set of samples. 

\subsection{Define datasets, object categories, and metadata }

Once images are have been collected, a dataset name, object categories,
and any additional metadata must be defined. Once this information
is recorded by the tool, a corresponding directory on the users filesystem
will be created to store dataset images. Images can be added to the
dataset by placing them into the directory. Labels can automatically
be applied to whole images by using descriptive directory names. For
example, a directory can be created for each of the stages of tomato
ripening, and images can be organized matching each stage. This process
can help improve efficiency by reducing the amount of meta-data manually
entered during the annotation process.

Using common object categories across datasets allows for the automatic
creation meta datasets. When exporting, images containing selected
categories can be grouped into larger or more specific datasets. 

\subsection{Segment instances and assign metadata}

The tool has provisions for adding multiple user accounts, which will
be associated with the annotations added to images. The super-user
defines the dataset, object categories, and metadata for others to
use. Users can be configured to start with an empty image and only
see their own annotations, or to begin with a copy of another users
result. This feature is intended to help increase efficiency and produce
more accurate labels through the combination of multiple results. 

In the dataset overview section, image are presented in a grid with
annotations overlaid and a note indicating the number of objects annotated.
Annotations can be added or modified by selecting an image.

On the annotation view, labels can be added using any of the previously
described techniques. A list of annotations is presented in a sidebar
which can be used to modify or delete individual labels. Hovering
the mouse cursor over each annotation reveals its label and metadata,
and double-clicking it brings up a dialog to edit those details. 

Work is automatically saved periodically and steps can be undone to
correct mistakes. Several per-user statistics are recorded during
annotation, including number of images annotated, average annotation
sizes, and time per annotation and per image.

\begin{figure}[H]
\begin{centering}
\includegraphics[width=0.9\columnwidth]{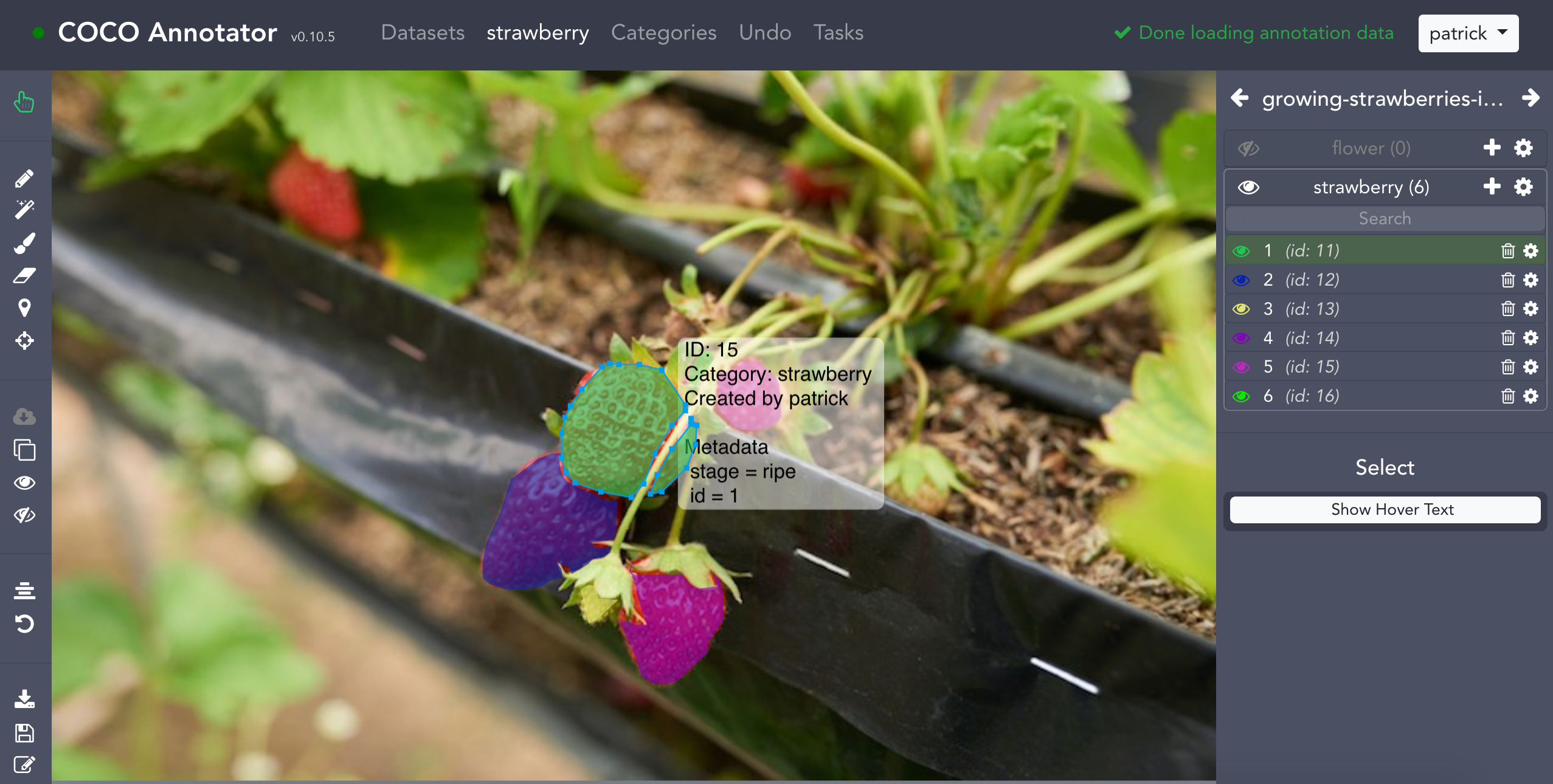}
\par\end{centering}
\caption{Image labeling tool user interface. Labeling techniques are chosen
using icons on the left. Annotated objects are listed in the right
sidebar. Hovering over an annotation displays its related metadata.
\label{fig:Image-labeling-tool}}

\end{figure}

\section{Dataset format}

File format conventions have played an important role in standardizing
and advancing the state of many digital systems. A common dataset
format can help facilitate and accelerate new applications and research
in agricultural image scene understanding. 

In order to integrate more closely with the general computer vision
community the dataset format chosen closely follows the \emph{common
objects in context }(COCO) dataset. COCO is an influential dataset,
used for pre-training models and as a benchmark for tracking the state-of-the-art.
Maintaining compatibility with COCO provides access to an extensive
array of existing support options for images, annotations, and for
evaluating results. 

The labeling tool generates annotation data in an extended COCO format,
adding features helpful for agricultural applications. The dataset
format uses JSON to store information about images and annotations
using attribute-value pairs. The extended data is stored under the
``poco'' (Plant Objects in COntext) attribute of each COCO data
structure. When used with native COCO tools the extra attributes simply
ignored and do not impacting original functionality. Figure \ref{fig:Extended-attributes-for}
shows the POCO dataset format. 

\begin{figure}[H]
\begin{centering}
\begin{lstlisting}[language=PHP,basicstyle={\small\ttfamily},breaklines=true,tabsize=2,comment={[l]{//}},commentstyle={\color{purple}\ttfamily},identifierstyle={\color{black}},keywords={typeof, new, true, false, catch, function, return, null, catch, switch, var, if, in, while, do, else, case, break},keywordstyle={\color{blue}\bfseries},morecomment={[s]{/*}{*/}},ndkeywords={class, export, boolean, throw, implements, import, this},ndkeywordstyle={\color{darkgray}\bfseries},sensitive=false,stringstyle={\color{MidnightBlue}\ttfamily}]
annotation {
  "id": int,
  "image_id": int,
  "category_id": int,
  "segmentation": RLE or[polygon],
  "bbox": [x, y, width, height],
  "keypoints" : [x1,y1,v1,...],
  ...
  "poco": {
    "maturity_stage": str,
    "plant_id": int,
    "keypoint_names" : [str],
    "skeleton" : [edge],
    ...
  }
} 
categories[{
  "id": int,
  "name": str,
  ...
  "poco": {
    "type": str,
  }
}]
\end{lstlisting}
\par\end{centering}
\caption{Extended attributes for object detection and key point tracking in
the COCO dataset format. All extended information is contained within
``poco'' (Plant Objects in COntext) attributes. In order to allow
for dynamic skeleton shape each annotation has its own associated
skeleton, instead of one defined for the entire category. \label{fig:Extended-attributes-for}}

\end{figure}

\section{Plant objects in context dataset}

Several thousand agricultural objects were annotated using the labeling
tool to create the initial series of POCO (Plant Objects in COntext)
datasets. Images were organized into subsets to provide finer-grained
download options for researchers. The three subsets created, corresponding
to different scene understanding and agricultural applications. 
\begin{enumerate}
\item Plant parts
\item Disease and pests
\item Plant development 
\end{enumerate}
The first subset contains images and annotations of a variety of plant
parts, including fruit, leaves, and stems. This subset currently focuses
on plants grown in greenhouses.

The second subset consists of images from the PlantVillage dataset
with annotations indicating the extent of disease on each leaf. This
dataset is aimed at better estimating leaf disease severity.

The third dataset contains time-lapse images of plants naturally developing.
The particular images in the initial dataset show tomato plants growing
in a commercial organic greenhouse over the course of 30 days. 

\begin{figure}[H]
\begin{centering}
\includegraphics[width=0.9\columnwidth]{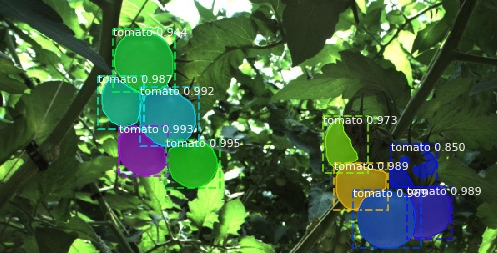}
\par\end{centering}
\caption{Plant parts. \label{fig:Plant-parts.}}
\end{figure}

\begin{figure}[H]
\begin{centering}
\includegraphics[width=0.9\columnwidth]{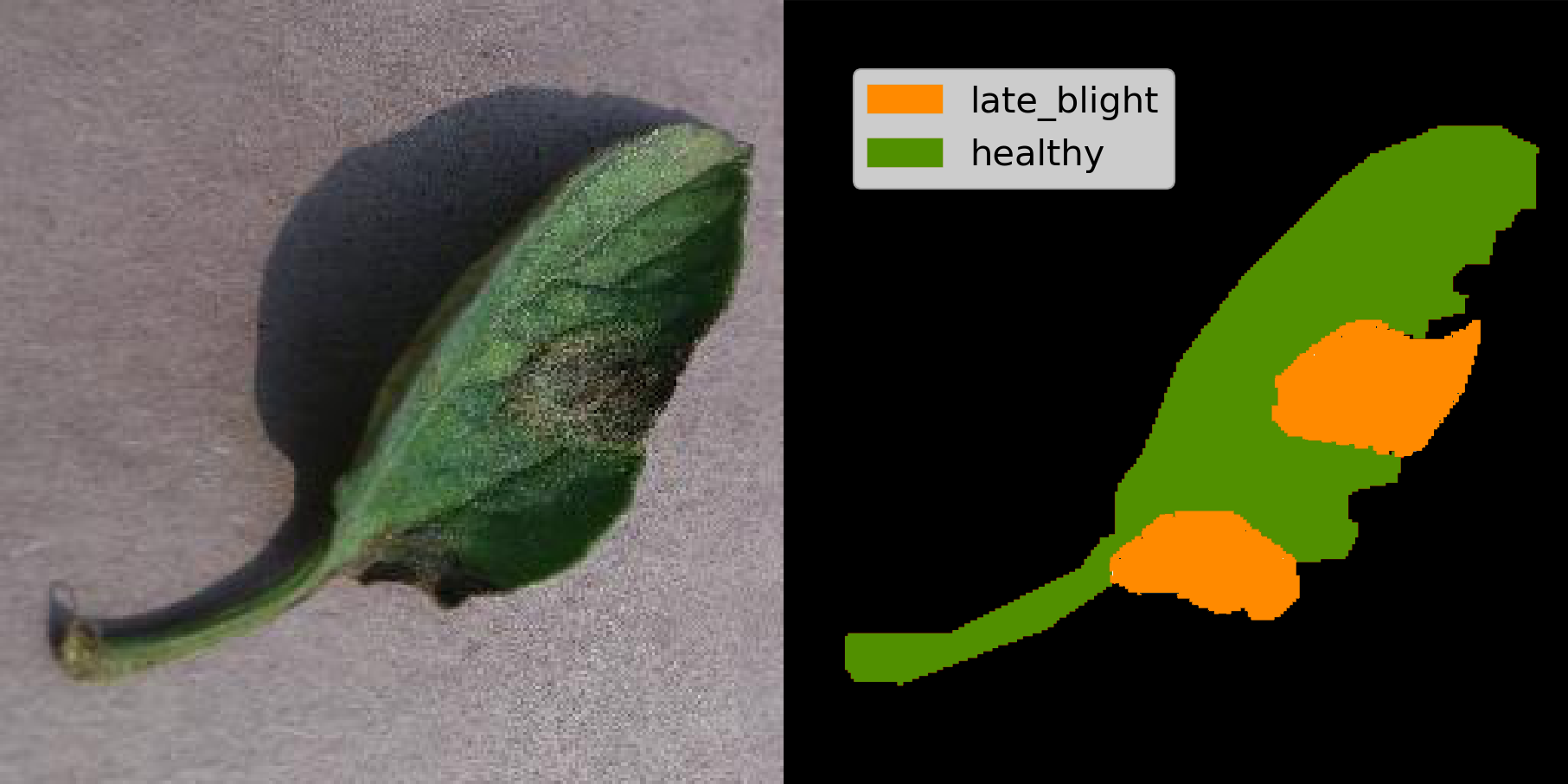}
\par\end{centering}
\caption{Disease. \label{fig:Disease.}}
\end{figure}

\begin{figure}[H]
\begin{centering}
\includegraphics[width=0.9\columnwidth]{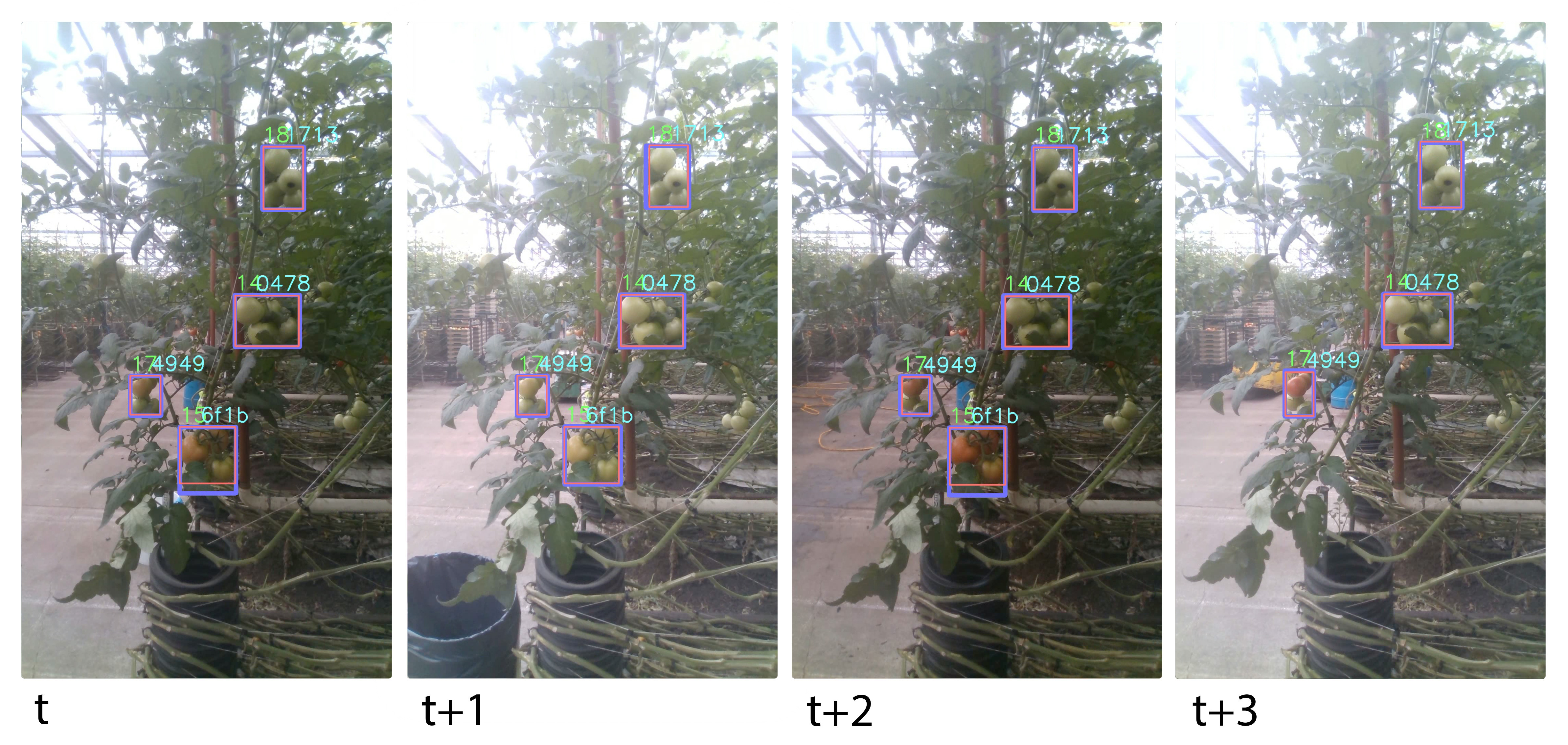}
\par\end{centering}
\caption{Time lapse. \label{fig:Time-lapse.}}
\end{figure}

\section{Conclusion}

In this paper we introduced a labeling tool multiple manual, semi-automatic,
and fully automatic annotation techniques, a new dataset format for
use in agricultural application, and an initial dataset with annotations
for plant parts, disease, and plant development. 

\bibliographystyle{plain}
\bibliography{label_paper,thesis}

\end{document}